\documentclass[10pt]{article}

\usepackage{framed,multirow}

\usepackage{amssymb}
\usepackage{latexsym}
\usepackage{tikz}
\usepackage{subfigure}
\usepackage{graphicx}
\usepackage{import}
\usepackage{adjustbox}
\usepackage{comment}
\usepackage{sidecap}
\usepackage{amsmath}
\usepackage{listofitems} % for \readlist to create arrays
\usepackage[outline]{contour} % glow around text
\usepackage{xcolor}
\usepackage{tabularx}

\usepackage{booktabs}

\usepackage{caption}
\usepackage{url}
\usepackage{xcolor}
\usepackage{multirow}
\usetikzlibrary{positioning}
\usetikzlibrary{matrix}
\usetikzlibrary{decorations}
\usetikzlibrary{arrows.meta} % for arrow size

\begin{document}

%\begin{frontmatter}


\begin{flushleft}
{\Large
\textbf\newline{CIMIL-CRC: a clinically-informed multiple instance learning framework for patient-level colorectal cancer molecular subtypes classification from H\&E stained images}}

Hadar Hezi\textsuperscript{1},
Matan Gelber\textsuperscript{2},
Alexander Balabanov\textsuperscript{2},
Yosef E. Maruvka\textsuperscript{3,4},
Moti Freiman\textsuperscript{1*}

%\\
\bigskip
\bigskip
\textbf{1} Faculty of Biomedical Engineering, Technion - Israel Institute of Technology, Haifa, Israel
\\
\textbf{2} Faculty of Electrical and Computer Engineering, Technion – Israel Institute of Technology, Haifa, Israel
\\
\textbf{3} Faculty of Biotechnology and Food Engineering, Technion - Israel Institute of Technology, Haifa, Israel
\\
\textbf{4} Lokey Center for Life Science and Engineering, Technion - Israel Institute of Technology, Haifa, Israel
\\
\bigskip
* moti.freiman@technion.ac.il
\end{flushleft}
\begin{abstract}
{\bf Background and Objective:} Treatment approaches for colorectal cancer (CRC) are highly dependent on the molecular subtype, as immunotherapy has shown efficacy in cases with microsatellite instability (MSI) but is ineffective for the microsatellite stable (MSS) subtype. There is promising potential in utilizing deep neural networks (DNNs) to automate the differentiation of CRC subtypes by analyzing hematoxylin and eosin (H\&E) stained whole-slide images (WSIs). Due to the extensive size of WSIs, multiple instance learning (MIL) techniques are typically explored. However, existing MIL methods focus on identifying the most representative image patches for classification, which may result in the loss of critical information. Additionally, these methods often overlook clinically relevant information, like the tendency for MSI class tumors to predominantly occur on the proximal (right side) colon. \\ 
{\bf Methods:} We introduce `CIMIL-CRC', a DNN framework that: 1) solves the MSI/MSS MIL problem by efficiently combining a pre-trained feature extraction model with principal component analysis (PCA) to aggregate information from all patches, and 2) integrates clinical priors, particularly the tumor location within the colon, into the model to enhance patient-level classification accuracy.
We assessed our CIMIL-CRC method using the average area under the receiver operating characteristic curve (AUROC) from a 5-fold cross-validation experimental setup for model development on the TCGA-CRC-DX cohort, contrasting it with a baseline patch-level classification, a MIL-only approach, and a clinically-informed patch-level classification approach. \\
{\bf Results:} Our CIMIL-CRC outperformed all methods (AUROC: $0.92\pm0.002$ (95\% CI 0.91-0.92), vs. $0.79\pm0.02$ (95\% CI 0.76-0.82), $0.86\pm0.01$ (95\% CI 0.85-0.88), and $0.87\pm0.01$ (95\% CI 0.86-0.88), respectively). The improvement was statistically significant. To the best of our knowledge, this is the best result achieved for MSI/MSS classification on this dataset. \\
{\bf Conclusion:} 
Our CIMIL-CRC method holds promise for offering insights into the key representations of histopathological images and suggests a straightforward implementation.
\end{abstract}


% \begin{highlights}
% \end{highlights}

%\begin{keyword}
%Multiple instance learning \sep Colorectal cancer \sep Digital pathology 
% %% MSC codes here, in the form: \MSC code \sep code
% %% or \MSC[2008] code \sep code (2000 is the default)
%\end{keyword}

%\end{frontmatter}

% \subsection{Entering text}
% \textcolor{newcolor}{\bf There is no page limit.}

\section{Introduction}
\label{sec:intro}
% importance of MSI/MSS classification
The notable effectiveness of immunotherapy in treating solid tumors like melanoma and lung cancer has spurred interest in applying this method to colorectal cancer (CRC) \cite{Ganesh2019-mw,Barzaman2021-cb}. However, the substantial variability both within and between CRC tumors complicates the success of immunotherapy \cite{molinari2018heterogeneity,Li2020-pt}. Notably, immune-checkpoint inhibitors have demonstrated encouraging results in CRC patients with microsatellite instability (MSI), highlighting the crucial role of molecular subtyping in CRC for tailoring treatment plans \cite{le2017mismatch, Hu2021PersonalizedStand}.

Polymerase chain reaction tests for DNA sequencing are currently the standard procedure for subtyping CRC \cite{Baudrin2018-tq}. However, accurate and precise polymerase chain reaction analysis requires an experienced team, specialized equipment, and is time-consuming, with limited sensitivity to samples with a low percentage of cancer cells. Consequently, MSI screening is often limited to high-volume tertiary care centers, making CRC subtyping into MSI and MSS, which is crucial for optimal management, less accessible to the broader population \cite{kim2021detection,Kather2019DeepCancer}. Faced with these challenges, convolutional neural networks (CNN) \cite{he2016deep, simonyan2014very} have emerged as a favored automated solution for identifying CRC subtypes using the readily available hematoxylin and eosin (H\&E) stained histopathological whole-slide images (WSI) \cite{Kather2019DeepCancer, kuntz2021gastrointestinal, wagner2023fully, Echle2020}. The primary benefit of this approach is its use of images already produced in routine clinical settings, offering an efficient and precise means for CRC subtyping that can easily be integrated into existing clinical processes.

While the vast information content in H\&E stained images is valuable for CRC subtyping, their extensive size, varying from 100 million to 10 billion pixels, poses a significant challenge to traditional graphics processing units during model training, primarily due to memory limitations. To address the challenge of handling large H\&E stained images, prior research has implemented a patch-based approach. This involves segmenting the images into smaller, more manageable sections for classification, followed by an aggregation of these results. For example, Kather et al. \cite{Kather2019DeepCancer} were at the forefront of using CNNs to determine CRC molecular subtypes (MSI and MSS) from H\&E images. They utilized a patch-level classification technique and combined the outcomes using a majority vote, achieving a moderate level of success on the TCGA-CRC-DX database (with an area under the receiver operating characteristic curve (AUROC) per patient of 0.77, n=360, 18\% MSI). Following a similar methodology, Echle et al. \cite{Echle2020} enhanced overall classification accuracy by expanding the dataset size, and integrating multiple databases. However, their performance on the TCGA-CRC-DX cohort did not surpass that of Kather et al. \cite{Kather2019DeepCancer}.

Yet, the classification of a WSI as either MSI or MSS pertains to the entire slide, not to individual patches. Consequently, only a fraction of the extracted patches may hold relevant information for this classification, with the remainder potentially being irrelevant. Therefore, a simplistic aggregation of patch-level classifications could potentially result in inaccuracies.

Multiple instance learning (MIL) frameworks are tailored for scenarios where each data sample is a `bag' comprising numerous smaller subsamples, but labeled collectively as a single unit \cite{carbonneau2018multiple}. In our context, each WSI is a `bag', constituted by its individual patches, and is classified as either MSI or MSS. MIL methodologies have been effectively utilized in managing extensive WSI datasets across a range of medical applications, including CRC WSI classification into MSI and MSS \cite{yao2020whole,zhang2022dtfd,shao2021transmil,sharma2021cluster,su2022attention2majority,qu2022dgmil,Bilal2021DevelopmentStudy, shats2022patient}.

MIL classification techniques based on embedded representations offer an alternative by forgoing the necessity for fixed-size data. They operate in a condensed dimensional space that captures the essence of the entire WSI. For example, Hemati et al. utilized k-means clustering on RGB histograms from WSI patches to form a compressed WSI representation \cite{Hemati2021CNNLearning}. They then selected random patch subsets from each cluster for WSI-level classification, using a consistent number of these embeddings. Fashi et al. enhanced this method by adding an attention mechanism to assign varying weights to different patches representing the WSI \cite{FASHI2022100133}. Similarly, Sharma et al. developed an end-to-end system that creates embeddings for each patch and uses k-means clustering for WSI classification, selecting a fixed number of vectors from each cluster \cite{sharma2021cluster}.

Despite these advancements, such approaches still necessitate selecting a certain number of patches or embeddings for final classification, risking information loss. Moreover, end-to-end training models require a substantial number of labeled WSIs, a challenging demand considering the typically small size of clinical WSI datasets \cite{zhang2022dtfd}.

Additionally, the scarcity of samples in these models limits their capability to generalize effectively across various tumor subtypes. Integrating specific clinical information could enhance their precision. Specifically, in CRC, tumors with the MSI classification are predominantly located in the ascending or proximal transverse colon (right-sided), while MSS tumors typically appear in the descending, sigmoid, and distal transverse colon (left-sided) \cite{Iacopetta2002-uq, sugai2006analysis}. Notably, within the TCGA-CRC-DX cohort referenced in this study, 87\% of MSI tumors are right-sided \cite{Kather2019DeepCancer}.
% The location of the tumor within the colon is usually given with the datasets. First in out data, TCGA and for example, in the datasets PLCO \cite{PLCO}, CRC-ICM-v1 \cite{CRC-ICM-v1} and more.

In this work, we introduce `CIMIL-CRC,' a clinically-informed MIL framework designed for classifying CRC WSIs as either MSI or MSS. The model employs a refined, task-specific feature extractor to map patch data into a latent space, after which these patch embeddings are consolidated using principal component analysis (PCA). We then select the eigenvectors with the largest eigenvalues, which capture features from the entire set of the patient’s patches. This approach provides a comprehensive representation of the data, avoiding the potential loss of information that might occur when focusing on only a few representative patches.
The classification at the WSI level is conducted by applying a classifier to the primary $k$ principal components, representing the entire WSI. Our method stands out by incorporating all patches from a WSI via PCA on their embeddings, rather than selecting $k$ key patches. We also improve the model's adaptability to diverse tumor subtypes by integrating known clinical information regarding the tumor's colonic location.

The `CIMIL-CRC' model has achieved, to our best knowledge, unparalleled MSI/MSS classification accuracy on the TCGA-CRC-DX test-set cohort \cite{Kather2019DeepCancer}.

The main contributions of our study are:

\begin{itemize}
\item Merging a pre-trained feature extraction model with PCA for MIL-based WSI classification.
\item Augmenting model accuracy through the inclusion of clinical prior knowledge.
\item Surpassing existing benchmarks in MSI/MSS subtype classification of CRC using H\&E stained WSI data from the TCGA-CRC-DX test-set cohort.
\end{itemize}

\section{Related work}

The standard DL-MIL algorithm for bag-level classification utilizes a variety of operators, including fully connected trainable layers and techniques like max-pooling, to derive the final WSI score. However,
these approaches face several limitations. Many rely on non-trainable, pre-defined pooling methods, such as the mean or maximum operators, to aggregate information from multiple instances within a bag. This restricts the flexibility of the model and can lead to suboptimal performance, as these fixed operators may not adapt well to complex datasets. Additionally, these methods struggle with interpretability since they do not indicate which instances within a bag contribute most significantly to the bag's label, a critical factor in applications like medical imaging where identifying key instances (e.g., regions of interest) is essential. To overcome this, Ilse et al. \cite{Ilse2018-co} proposed an attention mechanism that offers a trainable way to calculate a weighted sum of instances, thereby emphasizing crucial instances within the bag for WSI scoring. This mechanism also eliminates the need for bags to be of a fixed size. Bilal et al. \cite{Bilal2021DevelopmentStudy} applied an instance-level MIL strategy that focuses on identifying pivotal instances within a bag. Their `draw and rank' technique iteratively progresses the top 5 patches from each patient through training. For the test set, WSI-level classification was based on the average scores of these patches, resulting in an average AUROC of 0.9$\pm$0.01 and an area under the precision-recall curve (AUPRC) of 0.72$\pm$0.02 on the TCGA-CRC-DX database.

More recent efforts by Shats et al. \cite{shats2022patient} involved a preprocessing stage to create lower-dimensional embeddings for each bag, subsequently undergoing bag-level classification and aggregation to derive WSI-level results. However, these bag-level approaches necessitate the selection of a fixed number of patches for each classification batch, which can be problematic as the number of patches with relevant information for classification may vary among different cases.

Jiayun et al. \cite{li2021multi}, extracted suspicious regions for grade prediction in prostate histological WSI instead of relying on manual annotations. Their method consists of 2 stages. At the first stage, they classify bags of tiles as cancerous or benign. In the second stage, they choose discriminative tiles according to the attention weights from the previous stage to then classify to the cancer grade.  Alsaafin et al. \cite{alsaafin2023learning}, represented a WSI as a bag of instances and later on employed an attention and transformer mechanism to predict genes from the slide. 

Recent advancements have incorporated self-supervision into DL-MIL. 
% Our study employed traditional supervision for feature extraction, but we will reference these developments for result comparison since they utilized the same TCGA-CRC-DX database (n=360) as our study.
Schirris et al. \cite{schirris2022deepsmile} applied the self-supervision method simCLR \cite{chen2020simple} for feature extraction in their DeepSMILE framework. This process involves augmenting each patch repeatedly and training the network to align the latent spaces of the same patch while differentiating those of distinct patches. An innovation in the attention-based MIL field, VarMIL, calculates the variance within a bag's latent spaces and combines it with the attention MIL output before classification. The outcomes showed that simCLR alone (without MIL) achieved an AUROC of 0.87$\pm$0.01 and an F1-score of 0.61$\pm$0.11. When combined with VarMIL, the results were an AUROC of 0.86$\pm$0.02 and an F1-score of 0.47$\pm$0.1.

Leiby et al. \cite{leiby2022attention} used supervised feature extraction with VGG19 (pre-trained on ImageNet), along with a novel contrastive loss which compared the embeddings of each example to its bag-level feature aggregation calculated via attention rather than comparing it to an augmented patch's latent space. This method's results indicated that attention MIL alone yielded an AUROC of 0.86$\pm$0.01 and an AUPRC of 0.73$\pm$0.02 while adding contrastive loss improved the AUROC to 0.87$\pm$0.01 with an AUPRC of 0.69$\pm$0.04.

Saillard et al. \cite{saillard2021self} used self-supervision for feature extraction, combined with MIL models for patient aggregation. They employed momentum contrast v2 (MoCo v2 \cite{chen2020improved}) for feature extraction on the CRC dataset and utilized the DeepMIL \cite{Ilse2018-co} and the Chowder \cite{courtiol2018classification} MIL models for classification. The resulting AUROCs for these MIL models were 0.85$\pm$0.05 and 0.92$\pm$0.04 respectively.

Several recent studies have employed MIL frameworks for the analysis of WSIs, although not specifically for MSI/MSS classification in CRC. Lu et al. \cite{lu2021data} utilized attention-based learning to accurately classify whole slides. Their approach used a CNN to extract features from individual patches, representing each slide as a bag composed of these patches' feature vectors. The network, with attention mechanisms, highlighted the vectors with the highest diagnostic value, resulting in multi-class classification. Shao et al. \cite{shao2021transmil} introduced TransMIL, a MIL framework that captures correlations between instances within a bag using self-attention, treating instances as a sequence to more effectively model their relationships. Zhang et al. \cite{zhang2022dtfd} expanded the dataset by creating pseudo-bags, distributing instances from a single bag across three different pseudo-bags. Each pseudo-bag was processed through an attention-based MIL network, producing a vector representation that was fed into a second attention-based MIL network for final classification. 
Lu et al. \cite{lu2022slidegraph} developed SlideGraph+, a graph neural network based model, where each node stores a patch's feature vector, its coordinates, and its nuclear morphological features (extracted by HoverNet \cite{graham2019hover}). More recently, Fillioux et al. \cite{fillioux2023structured} proposed structured state space models to optimally project WSIs into compressed representations for classification, and Shi et al. \cite{shi2024masked} introduced the masked hypergraph learning framework for weakly supervised WSI classification.

Despite the advancements, these models primarily focus on WSI images and often overlook the potential benefits of incorporating additional clinical information that could improve classification accuracy. Recent studies have highlighted the value of integrating clinical data with imaging in developing machine-learning models for clinical applications. For instance, Gilad et al. combined patient clinical features with diffusion-weighted MRI data to predict responses to Neoadjuvant chemotherapy in breast cancer patients \cite{gilad2022pd}. Similarly, Guez et al. \cite{GUEZ2022107207} integrated features from magnetic resonance enterography with biochemical biomarkers to predict Crohn's disease status in the terminal ileum.

\section{Methods}
\label{sec:methods}
 \begin{figure*}
 \centering
  %\centerline{
   \import{./}{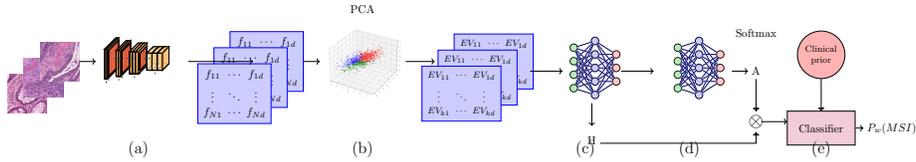}
   %}
   
    \caption[CIMIL-CRC architecture]{ \textbf{Model architecture overview:} (a) Patches from each patient are processed through a pre-trained Efficient-net, operating in evaluation mode, to perform patch classification. This yields extracted features of dimension [2560,7,7]. These feature sets for each patient are then arranged into matrices and stored. (Image of CNN by Haris Iqbal \cite{cnn_im}). (b) PCA is subsequently applied to each patient’s feature matrix, retaining and saving the top-k eigenvectors. (c) These eigenvector matrices are input into an MLP for further feature extraction, resulting in the matrix $H \in \mathcal{R}^{kx512}$ (MLP illustration by Izaak Neutelings \cite{MLP_im}). (d) Another MLP processes matrix $H$ to create the attention matrix $A \in \mathcal{R}^{kx3}$, followed by applying the $softmax$ function to its rows. The matrices $A$ and $H$ are then combined through multiplication. (e) The combined output is flattened and introduced into a classifier MLP to determine the patient's MSI score. This MSI score is subsequently adjusted by the side function (Equation~\ref{eq:side}), generating the final patient-specific MSI score.}
  \label{fig:evmil_arch}
\end{figure*}

The comprehensive architecture of our CIMIL-CRC approach is depicted in Figure~\ref{fig:evmil_arch}. The primary steps of our technique are as follows:

\begin{itemize}
\item \textbf{Feature Extraction (Figure~\ref{fig:evmil_arch} a):} In the initial phase, all patches are processed for feature extraction using a baseline model that has been pre-trained. The features derived from this stage encapsulate vital visual data, forming the basis for further analysis. These features are then systematically compiled into matrices specific to each patient.
\item \textbf{Embedding Representation (Figure~\ref{fig:evmil_arch} b):} Using the patch-level features, we apply PCA to construct a new patient-centric representation. This step effectively simplifies the feature space while maintaining key elements unique to each patient, thus optimizing computational efficiency without compromising on critical information.
\item \textbf{Patient-level Classification (Figure~\ref{fig:evmil_arch} c-e):} Moving forward, the simplified feature representation facilitates patient-level classification via a multi-layer perceptron (MLP). The integration of an attention mechanism within our MLP enables it to synthesize insights from diverse patient components, leading to more comprehensive and precise patient-level assessments.
\item \textbf{Clinical Data Incorporation (Figure~\ref{fig:evmil_arch} e):}
We finally account for the significant difference in the anatomical location of MSI and MSS tumors in the colon by adjusting the classifier output by a predefined factor (Equation~\ref{eq:side}). This refinement boosts the efficacy of our model, culminating in the development of the CIMIL-CRC framework.
\end{itemize}

\subsection{Feature extraction}
We first implemented a baseline CNN model for patch-level binary classification (MSI/MSS) using a transfer-learning approach. To construct this model, we leveraged the pre-trained Efficient-net b7 architecture \cite{Tan2019-fb}, which had been originally trained on the extensive Imagenet database \cite{deng2009imagenet}. Fine-tuning was performed by retraining the last 7 blocks of layers, along with the fully connected layers, using our dataset. We saved a checkpoint of the trained baseline model. 

We then set the system to evaluation mode to extract features from every patch in both the training and test sets, utilizing the penultimate layer of the EfficientNet architecture. The extracted features, with dimensions \([C, H, W] = [2560, 7, 7]\) are directly derived from EfficientNet, providing a detailed and information-rich representation.

For enhanced higher-level analysis, we organize the extracted features for each patient into a matrix, denoted as $F_w$: 
\begin{equation}
F_w =
\begin{pmatrix}
f_{11} & \cdots & f_{1d}\\
\vdots & \ddots & \vdots \\
f_{N1} & \cdots & f_{Nd}
\end{pmatrix} 
\end{equation}
Here, for a patient with $N$ patches, the matrix $F_w$ has dimensions $N \times d$, where $d = (7 \times 7 \times 2560)$. Each row in this matrix corresponds to a flattened feature vector $f_i$, which is derived from the $i$-th patch.

% For enhanced higher-level examination, we arrange the gathered features for each patient into a matrix, which we denote as $F_w$:
% \begin{equation}
% F_w =
% \begin{pmatrix}
% f_{11} & \cdots & f_{1d}\\
% \vdots & \ddots & \vdots \\
% f_{N1} & \cdots & f_{Nd}
% \end{pmatrix} 
% \end{equation}

% In the case of a patient with $N$ patches, this matrix assumes the dimensions of $N\times d$, where $d$ is calculated as $(7 \times 7 \times 2560)$. Every row in this matrix represents a flattened feature vector $f_{i}$ derived from the $i-th$ patch:

The number of patches varies from one patient to another, and accordingly, their feature matrices are saved. By structuring the features in this format, we efficiently consolidate and arrange the patch-level data into a format centered around each patient. This setup enables us to delve into patient-specific patterns and correlations, fostering deeper and more holistic analyses of the data.

\subsection{Embedded representation of the extracted feature matrices}
To facilitate patient-level classification, we sought a more condensed representation of our feature matrices. This approach enables us to process an entire patient's data in a single instance and subsequently generate the network's score for that patient.

To achieve this, we implemented the PCA method for a new, comprehensive data representation. This method also ensures the encapsulation of the entire original data set rather than selecting representative patches as suggested by previous MIL approaches.

For each patient, the extracted eigenvectors define new axes in their feature space, capturing the variance among features. To accommodate varying dimensionality, we selected the top-$k$ eigenvectors based on their corresponding eigenvalues, forming a matrix for each patient $E_w \in \mathcal{R}^{k \times d}$. Since the number of eigenvectors is constrained by the smaller value between the number of samples and features, the patients with fewer patches due to Kather et al.'s \cite{Kather2019DeepCancer} random patch sampling for balancing MSS patient data, may yield fewer eigenvectors. In these cases, we included as many eigenvectors as possible, ensuring the patient data remained part of the analysis.

% For each patient, the extracted eigenvectors delineate new axes in their feature space, effectively encapsulating the variance among features. To handle varying dimensionality, we chose the top-k eigenvectors based on their eigenvalues, creating a new matrix for each patient $E_w \in \mathcal{R}^{k \times d}$. Since the maximum number of eigenvectors that can be derived from the feature matrix is limited by the smaller value between the number of samples and the number of features. As a result, for certain patients, particularly those with fewer patches due to the random patch sampling method by Kather et al. \cite{Kather2019DeepCancer} used for balancing patch-level data in MSS patients, fewer eigenvectors may be obtained. In these cases, instead of excluding the patients from the analysis, we used the highest number of eigenvectors that could be derived from the available data.

% Nonetheless, considering that the maximum number of eigenvectors that can be derived from the feature matrix is constrained by $MIN(n_{samples}, n_{features})$, there are instances where patients, particularly those with a smaller number of patches due to Kather et al.'s \cite{Kather2019DeepCancer} method of random patch sampling in MSS patients for patch-level data balancing in the training set, may yield fewer eigenvectors. In such cases, we utilized the maximum possible number of eigenvectors for these patients instead of excluding them from the analysis.

\subsection{Patient-level classification} 
We then used the selected eigenvectors as input to the DeepMIL with attention \cite{Ilse2018-co} classifier to obtain the patient-level classification as follows. The matrix of each patient's $k$  eigenvectors, $E_w$ undergoes a feature extraction process through an MLP, resulting in $H = F(E_w), \ H\in \mathcal{R}^{k \times 512}$. 
The attention matrix is computed as follows: \begin{equation} A=W_2\cdot tanh(W_1 \cdot H) \odot W_3 \cdot sigmoid(W_4 \cdot H); \ A \in \mathcal{R}^{k\times 3} \end{equation} where $W_i$, $i \in {1, \dots, 4}$ are the learned weights of the network, and $\odot$ represents element-wise multiplication. The dimension of the attention matrix $A$ is $k \times 3$, corresponding to 3 rows of weights generated from the eigenvectors in $H$.

Next, we apply the $softmax$ function across the rows of $A$ to normalize the weights. The final probability output for a patient's MSI status is calculated as: \begin{equation} P_w(\text{MSI}) = sigmoid(W_5 \cdot A^T \cdot H) \end{equation} where $W_5$ are the learned weights.

\subsubsection{Clinical Data Incorporation}
Finally, we incorporated the clinical prior, given with our data, associated with the anatomical side of the patient's tumor by multiplying the classifier's output $P_w(\text{MSI})$ by a predefined factor:
\begin{equation}
P_w(\text{MSI})_{final} = P_w(\text{MSI}) \cdot Prior
\end{equation}
where $Prior$ reflects the prevalence of an MSI tumor given the clinical information. Specifically, as indicated by prior research \cite{Iacopetta2002-uq,sugai2006analysis} and consistent with the data trends in the TCGA-CRC-DX cohort referenced in this study \cite{Kather2019DeepCancer}, about 90\% of MSI tumors are found on the right side. We therefore defined the side prior function based on the patient's tumor's location as:
\begin{equation}
Prior = f(side) = \left\{
        \begin{array}{ll}
             0.1 & \quad side = left \\
            \beta & \quad side = right/undefined
        \end{array}
    \right.
   \label{eq:side}
\end{equation}
All patients' scores undergo multiplication by this factor. If the tumor is located on the left side, the classifier's output for the patient is scaled by a factor of 0.1 which is equivalent to the left-side prevalence. For tumors on the right side or those with an undefined side, the classifier output was tested with several values on the validation set.

\subsection{Training settings}
We used the binary cross entropy loss with label smoothing \cite{DBLP:journals/corr/SzegedyVISW15} as our loss function to tackle the over-fitting and over-confidence of the classifier.
% move the alpha range to results or experiments
%was tested with values of $\alpha=[0,0.01,0.02,0.03]$.
We accounted for data imbalance in the training dataset by incorporating class-related weights into the loss function. We calculated the class weights as the inverse proportion of each patient's class representation.

We employed the ADAM optimizer for training with the following parameters: betas were set to (0.7, 0.99), an initial learning rate of 0.0001, a training duration of 10 epochs, and a batch size configured for one patient. Model checkpoints were saved whenever MSI accuracy and overall accuracy exceeded 95\%. The cross-validation folds were stratified, aligning with the baseline methodology. We implemented our code using pyTorch 1.9 and ran it on Nvidia A100 graphics processing units using a PyTorch NGC Container, version 21.04.

\section{Experiments}
\label{experiments}

\subsection{Dataset}
\label{data}
Our research utilized the TCGA-CRC-DX cohort from the cancer genome atlas \cite{crc-data}, which comprises data from n=360 patients. This cohort consists of formalin-fixed paraffin-embedded diagnostic slides that are H\&E stained from CRC patients. Additionally, the database provides results of DNA mutations, RNA expression, the anatomic site of the tumor within the colon, and more. MSI/MSS reference labels were determined by \cite{Liu2018ComparativeAdenocarcinomas} (as referenced in Supplementary Table 2 of \cite{Kather2019DeepCancer}). Kather et al. pre-processed and published this data, with the steps detailed in \cite{Kather2019DeepCancer}.

\begin{figure}[t!]
    %\centering
        %\documentclass[border=15pt, multi, tikz]{standalone}
%\usepackage{import}
%\usetikzlibrary{positioning,matrix}
%\usetikzlibrary{shapes}
%\usetikzlibrary{plotmarks}
%\usetikzlibrary{decorations.pathreplacing,calligraphy}
%\begin{document}
\begin{adjustbox}{width=\linewidth}  
\begin{tikzpicture}[font=\normalsize,thick]
% Start block
\node[draw,
    minimum width=2.5cm,
    minimum height=1cm] (block2) {Entire Cohort - n=430};

% Power and voltage variation
\node[draw,
    align=center,
    below=of block2,
    minimum width=3.5cm,
    minimum height=1cm
] (block3) {Pre-processed by Kather et al. - n=360};

% Conditions test
\node[draw,
    align=center,
    circle,
    below right=of block3,
    minimum width=2.5cm,
    text width=2cm,
    inner sep=0,fill=red!30] (block4) {Test - n=100, p=98,904};
    
\node[draw,
    align=center,
    circle,
    below left=of block3,
    minimum width=2.5cm,
    text width=2cm,
    inner sep=0,fill=red!30] (block5) {Train - n=260, p=93,408};

\node[draw=black,
    below left=of block4,
    minimum width=2.5cm,
    text width=2.5cm,
    minimum height=1cm, fill=green!30] (block6) 
    { MSI: \\ n=26, p=28,335};

\node[draw=black,
    below right=of block4,
    minimum width=2.5cm,
    text width=2.7cm,
    minimum height=1cm,fill=green!30] (block7) 
    { MSS: \\ n=74, p=70,569};

\node[draw=black,
    below left=of block5,
    minimum width=2.5cm,
    text width=2.5cm,
    minimum height=1cm, fill=green!30] (block8) 
    { MSI: \\ n=39, p=46,704};

\node[draw=black,
    below right=of block5,
    minimum width=2.5cm,
    text width=2.5cm,
    minimum height=1cm,fill=green!30] (block9)  { MSS: \\ n=221, p=46,704};

% Arrows
\draw[-latex] 
    (block2) edge (block3)
    (block3) edge (block4);

\draw[-latex] (block3) -- (block5)
    node[pos=0.25,fill=white,inner sep=0]{};
\draw[-latex] (block4) -| (block6)
 node[pos=0.25,fill=white,inner sep=0]{};
    
\draw[-latex] (block4) -| (block7)
    node[pos=0.25,fill=white,inner sep=0]{};
\draw[-latex] (block5) -| (block8)
 node[pos=0.25,fill=white,inner sep=0]{};
    
\draw[-latex] (block5) -| (block9)
    node[pos=0.25,fill=white,inner sep=0]{};

\end{tikzpicture}

\end{adjustbox}
%\end{document}

\caption{Data preprocessing chart. From the TCGA-CRC cohort (n=430), Kather et al. selected and pre-processed a subset (n=360) for publication. This subset was divided into a training set (n=260) and a testing set (n=100). Balancing was applied at the patch level within the training set by excluding MSS patches.}
\label{fig:data_split}
\end{figure}
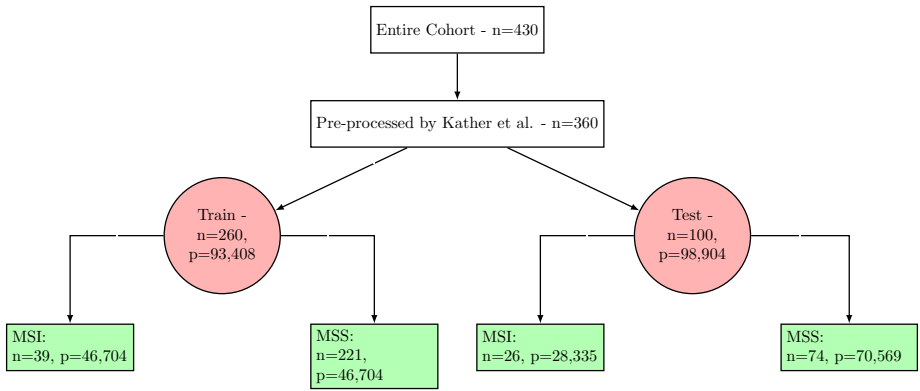

Figure \ref{fig:data_split} illustrates the distribution of patients in the dataset. During the pre-processing phase, a random selection of 100 out of the 360 patients was set aside as a test set \cite{Kather2019DeepCancer}. In the training set, 15\% of patients are classified as MSI, while the test set contains 26\% MSI patients. Multiple image patches were extracted from each H\&E slide, and the training set was balanced at the patch level by selectively discarding MSS patches.

\subsection{Experimental methodology}

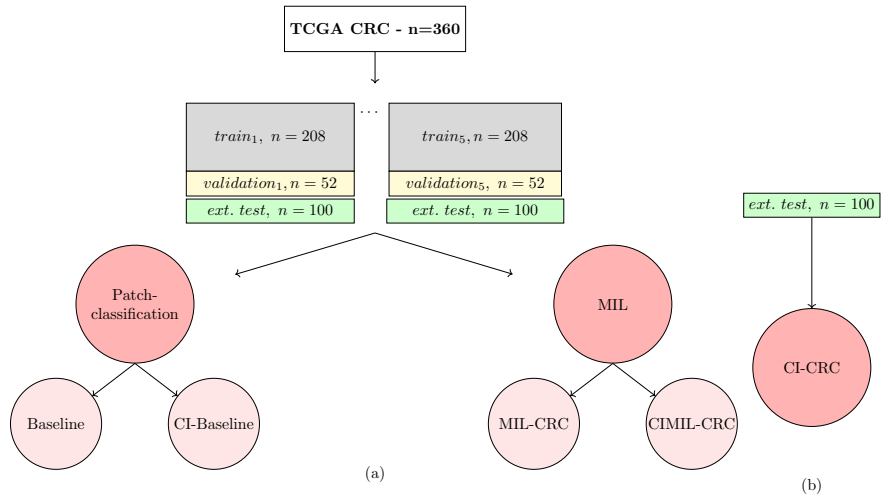
\begin{figure}[t!]
    %\centering
    \adjustbox{max width=\textwidth} {
    %\begin{tabular}{cc}
   \def \basebp[#1]#2#3{
%\begin{scope}[shift={#2}]
\node[draw,
    align=center,
    xshift=0.5cm,
    below =1cm of #1,
    minimum width=1cm,
    minimum height=1cm
] (block#2) { \textbf{Baseline}};
\node[draw,
    align=center,
    right=0.5cm of block#2,
    minimum width=1cm,
    minimum height=1cm
] (block#3) { \textbf{BP-CNN}};
%\end{scope}
}

\def\m#1#2[#3]{
\matrix [ampersand replacement=\&,below= of #3]{
\node[draw,
    align=center,
    minimum width=3.7cm,
    minimum height=1.5cm,
    fill=gray!30
] (block#1) {$train_1, \ n=208$};
\node[draw,
    align=center,
    below=-\pgflinewidth of block#1,
    minimum width=3.7cm,
    minimum height=0.5cm,
    fill=yellow!20
] {$validation_1, n=52$};
\node[draw,
    align=center,
    below=0.6cm of block#1,
    minimum width=3.7cm,
    minimum height=0.5cm,
    fill=green!20
]  {$ext. \ test, \ n=100$};\&
\node {$\cdots$}; \&
\node[draw,
    align=center,
    minimum width=3.8cm,
    minimum height=1.5cm,
    fill=gray!30
] (block#2) {$train_5, n=208$};
\node[draw,
    align=center,
    below=-\pgflinewidth of block#2,
    minimum width=3.8cm,
    minimum height=0.5cm,
    fill=yellow!20
]  {$validation_5, \ n=52$};
\node[draw,
    align=center,
    below=0.6cm of block#2,
    minimum width=3.9cm,
    minimum height=0.5cm,
    fill=green!20
]  {$ext. \ test, \ n=100$};\\ };
}

\begin{adjustbox}{width=\linewidth}   
\begin{tikzpicture}

\node[draw,
    align=center,
    minimum width=3.5cm,
    minimum height=1cm
](block1) at (0,0)  { \textbf{TCGA CRC - n=360}};
%train-validation 
\m{5}{6}[block1]; 
\node[draw,
    align=center,
    circle,
    %xshift = -3cm,
    %node distance = 3cm,
    below left =2cm and 0.2cm of block5,
    minimum width=2.6cm,
    text width=2cm,
    inner sep=0,fill=red!30] (block2) {Patch-classification};
\node[draw,
    align=center,
    circle,
    %xshift = -3cm,
    %node distance = 0.5cm,
    below left=1cm and 0.1cm of block2,
    minimum width=1.8cm,
    text width=2cm,
    inner sep=0,fill=red!10] (Baseline) {Baseline};
\node[draw,
    align=center,
    circle,
    %xshift = -3cm,
    %node distance = 0.5cm,
    below right=1cm and 0.1cm of block2,
    minimum width=1.8cm,
    text width=2cm,
    inner sep=0,fill=red!10] (CI-Baseline) {CI-Baseline};
%\draw[->]    (block2) --(-13.1,-7);

 %\end{scope}
\node[draw,
    align=center,
    circle,
    %node distance=5cm,
    %right = 3cm of block2,  
    below right = 2cm and 0.2cm of block6,
    minimum width=2.6cm,
    text width=2cm,
    inner sep=0,fill=red!30] (block3) {MIL};
\node[draw,
    align=center,
    circle,
    %node distance=0.1cm,
    below left=1cm and 0.1cm of block3,  
    minimum width=1.8cm,
    text width=2cm,
    inner sep=0,fill=red!10] (MIL-CRC) {MIL-CRC};
    \node[draw,
    align=center,
    circle,
    %node distance=0.1cm,
    below right= 1cm and 0.1cm of block3,  
    minimum width=1.8cm,
    text width=2cm,
    inner sep=0,fill=red!10] (CIMIL-CRC) {CIMIL-CRC};
  %  \draw[->]    (block3) --(-2.2,-7);
   \node [below = of block1, yshift = -8cm] (a) {(a)};   

\node[draw,
    align=center,
    %below=0.5cm of block#1,
    right= of block6,
    xshift=3cm,
    yshift = -1.5cm,
    minimum width=3cm,
    minimum height=0.5cm,
    fill=green!20
]  (test) {$ext. \ test, \ n=100$};

\node[draw,
    align=center,
    circle,
    %xshift = -3cm,
    %node distance = 3cm,
    below =2cm and 0.2cm of test,
    minimum width=2.6cm,
    text width=2cm,
    inner sep=0,fill=red!30] (CI-CRC) {CI-CRC};
 \node [below = of CI-CRC] (b) {(b)};

% Arrows
%\draw[->]    (-12.9,-10.5) --(block7);
%\draw[->]    (-12.9,-10.5) --(block8);
\draw[->](block1) -- (0,-1.2);
\draw[->, shorten >=1cm](0,-4.5) -- (block2);
\draw[->, shorten >=1cm](0,-4.5) -- (block3);
%\draw[->, shorten <=1cm] (block5.south) -- (block2);
%\draw[->, shorten <=1cm] (block6.south) -- (block3);
\draw[->] (block2.south) -- (Baseline);
\draw[->] (block2.south) -- (CI-Baseline);
\draw[->] (block3.south) -- (MIL-CRC);
\draw[->] (block3.south) -- (CIMIL-CRC);
\draw[->] (test) -- (CI-CRC);
%\draw[-latex] 
%    (block1) edge (block2)
%    (block1) edge (block3);
    % (block2) edge (block5)
    % (block2) edge (block6)

\end{tikzpicture}
\end{adjustbox}
    %\import{./}{images/CI-CRC_experiment.tex}
    }
   %\end{tabular} 
    \caption[experimental flow]{
    (a) Our external test set consists of n=100 patients, randomly selected by Kather et al. from their total cohort of n=360 \cite{Kather2019DeepCancer}. The remaining n=260 patients were meticulously divided into 5 stratified folds to enable cross-validation, ensuring a balanced class distribution across these folds.
    We evaluated the efficacy of our methods using the AUROC, AP, F1-score, and Cohen's kappa scores, applied to each fold's model on the external test set. This evaluation was conducted for both the patch classification models and our MIL-based models (MIL-CRC, CIMIL-CRC). Additionally, for a comprehensive comparison, we incorporated the clinical information prior also to the patch-classification model (resulting in CI-Baseline). (b) The setup for using clinical side information only as the classification parameter. As this experiment is deterministic we applied it to the test set only (n=100) and reported Cohen's kappa and accuracy. }
  \label{fig:experiments}
\end{figure}

Figure~\ref{fig:experiments} provides an overview of our experimental approach. Initially, we detail the various models deployed in our assessment, followed by a description of our evaluation methodology.

\subsubsection{Classification models}

We evaluated the the distinct contribution of each component in our CIMIL-CRC methodology, by conducting comparative analyses with various models in addition to our CIMIL-CRC model as follows:

\begin{itemize}
\item \textbf{Baseline Model:} This model served as our point of reference. It primarily focused on patch-level classification with patient aggregation conducted externally to the CNN.
\item \textbf{CI-Baseline - clinical information enhanced baseline model:}
This version enhances the baseline model by incorporating clinical side information during patient aggregation, allowing us to assess its contribution to overall model performance in comparison to other approaches.
\item \textbf{CI-CRC - clinical side information only}
To determine the usefulness of the CNN, we conducted an experiment using only clinical side information to classify patients to MSI/MSS.
\item \textbf{MIL-CRC models}
Our main model utilizes our PCA-based MIL approach (MIL-CRC) with and without adjusting the model output with the side prior function.
\end{itemize}

\subsubsection*{Baseline model}
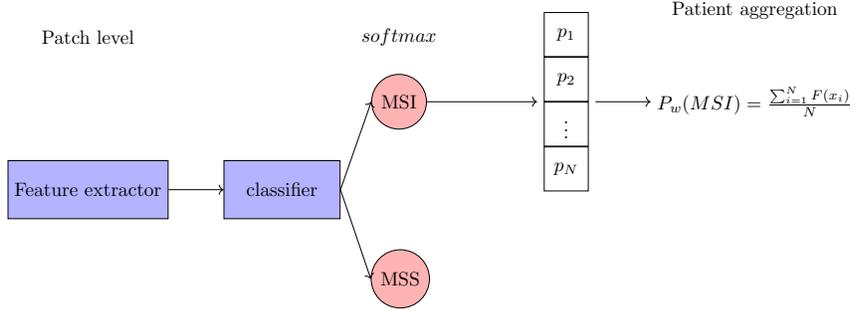
\begin{figure}[t!]
    %\centering
    \resizebox{\columnwidth}{!}{
   %\usetikzlibrary{matrix}
%\usetikzlibrary{positioning, arrows.meta}
%\usetikzlibrary{shapes,arrows}

\tikzstyle{block} = [draw, rectangle, 
    minimum height=3em, minimum width=6em,fill=blue!30]
\tikzstyle{circ} = [draw, circle, node distance=1cm,fill=red!30]
\tikzstyle{input} = [coordinate]
%\begin{figure}
\resizebox{9.2cm}{!}{%
\begin{tikzpicture}
\node [block] (inception) {Feature extractor};
\node [block, right = 1cm of inception] (mlp) {classifier};
\node [above = 2cm of inception] {Patch level};
\node [circ, above right=of mlp] (msi) {MSI};
\node [above=0.4cm of msi]{$softmax$};
\node [circ, below right=of mlp] (mss) {MSS};

\matrix (A) [matrix of math nodes,right=2cm of msi, nodes={draw, minimum size=8mm}, column sep=-\linewidth]
{ p_1\\p_2\\\vdots\\p_N\\};
\node(B)[right=of A]{
    \(P_w(MSI) = \frac{\sum_{i=1}^{N}F(x_i)}{N}\)};
\node [above=of B]{Patient aggregation};
\draw [->] (inception.east) -- (mlp.west);
\draw [->] (mlp.east) -- (msi.west);
\draw [->] (mlp.east) -- (mss.west);
\draw [->] (msi.east) -- (A.west);
\draw[->] (A.east) -- (B.west);
\end{tikzpicture}
}
%\end{figure}
}
    \caption[Baseline Model Architecture]{Baseline model architecture. Patches are fed into the Efficient-net b7 for feature extraction. The last two layers consist of fully connected classifier layers. The outputs then pass through a $softmax$ layer to generate probabilities. $N$ represents the number of patches for each patient, with $F(x_i)$ denoting the MSI probabilities of these patches. The MSI score for each patient, $P_w$, is calculated as the average of these MSI probabilities.}
  \label{fig:base_arch}
\end{figure}

Figure \ref{fig:base_arch} illustrates the baseline model architecture employed in our research. It utilizes an Efficient-net b7 as the CNN to classify patches into MSI/MSS categories. Patch-level classifications are combined into a patient-level classification by averaging the probabilities of patch classification. Specifically, for a classifier $F$, the output for a given patch $x$ represents the probability that it belongs to the MSI class, with $\left(0 \leq F(x) \leq 1\right)$.

For an H\&E slide $W$, $N$ patches $(x_1, x_2, \hdots, x_N)$ are extracted from $W$. The MSI probability at the patient level is then calculated as follows:
\begin{equation}
P_w(\text{MSI}) = \frac{\sum_{i=1}^{N}F(x_i)}{N}
\label{eq:agg}
\end{equation}

\subsubsection*{CI-Baseline - Baseline model with clinical side Information}
A baseline patch-level classification model employing a consolidated score for patient-level classification, as defined in Equation~\ref{eq:agg}, complemented by the integration of clinical side information. This integration involves multiplying the patient score (Equation~\ref{eq:agg}) with the side prior function (Equation~\ref{eq:side}).

\subsubsection*{CI-CRC - Clinical Side Information only}
This method classified patients to MSI/MSS using their clinical side information only.
If a patient is right-sided then its predicted label will be MSI, if it is left-sided then its predicted label will be MSS. 
Interestingly, the probability in our dataset to be MSI given the side of the tumor is given by the Bayes rule:
\[P(MSI|right) = \frac{P(right|MSI)P(MSI)}{P(right)} = \frac{0.87\times 0.18}{0.44} = 0.35\]
\[P(MSI|left) =  \frac{P(left|MSI)P(MSI)}{P(left)} = \frac{0.13\times 0.18}{0.56}=0.04 \]
 This experiment doesn't have a AUROC or AUPRC as the prediction is deterministic based on the side information.
\subsubsection*{MIL-CRC models}
Our main PCA-based MIL model, with and without integrating clinical side information as a prior.

\subsubsection{Evaluation methodology}
We started with hyperparameters optimization. Specifically, we consider the optimal number of eigenvectors necessary for accurately representing a patient and the impact of different label smoothing rates on the classification performance. 
We determined the optimal number of eigenvectors to use by utilizing the CIMIL-CRC model with a varying number (range: $[1,120]$) of eigenvectors as input. For each specified number of eigenvectors, we conducted a stratified 5-fold cross-validation, followed by an analysis of the average F1-score and Cohen's kappa results on the validation set across the folds, considering the number of eigenvectors as input to the final classifier. Additionally, we tested smoothing rates of $[0, 0.01, 0.02, 0.03]$.

Subsequently, we trained the various models (Baseline, CI-Baseline, MIL-CRC, CIMIL-CRC) using a stratified 5-fold cross-validation method, resulting in five trained models for each approach.

In the final phase, we compared these models' MSI/MSS classification accuracy on the test set provided by Kather et al. \cite{Kather2019DeepCancer}. We used the AUROC, the AUPRC, the F1-score, the Cohen's kappa score, and the accuracy as our evaluation metrics. To compare the performance of the Baseline models against the MIL-CRC models, we applied a paired Student's t-test to the AUROC and AUPRC results, and a McNemar's test to the final predictions, using a significance level of p$<$0.05 to determine the statistical significance of the improvements observed with MIL-CRC and CIMIL-CRC compared to the Baseline models.

\section{Results}
\begin{figure}[t!]
    \centering
    \subfigure[]{\includegraphics[width=0.45\textwidth]{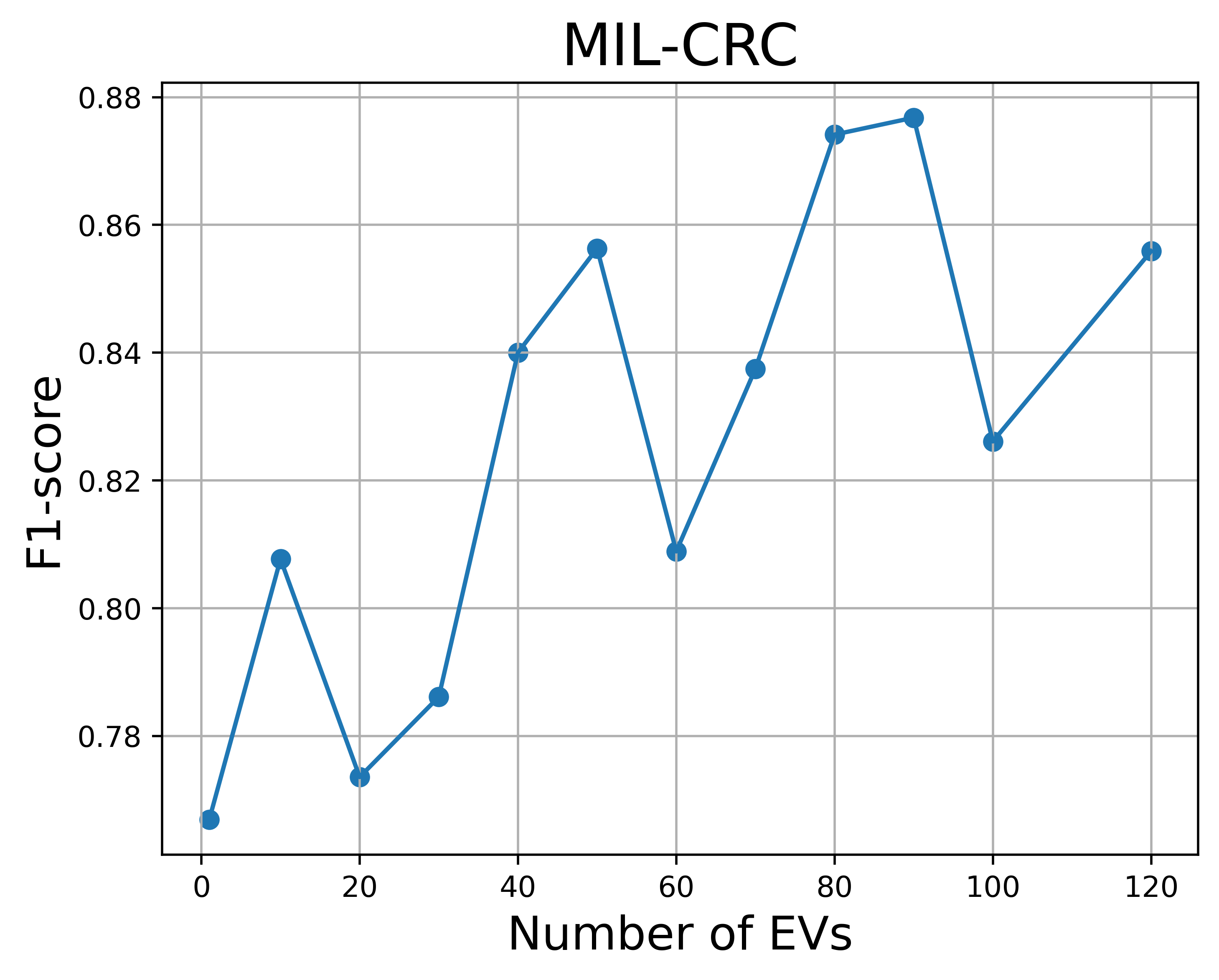}}
    \hfill
    \subfigure[]{\includegraphics[width=0.45\textwidth]{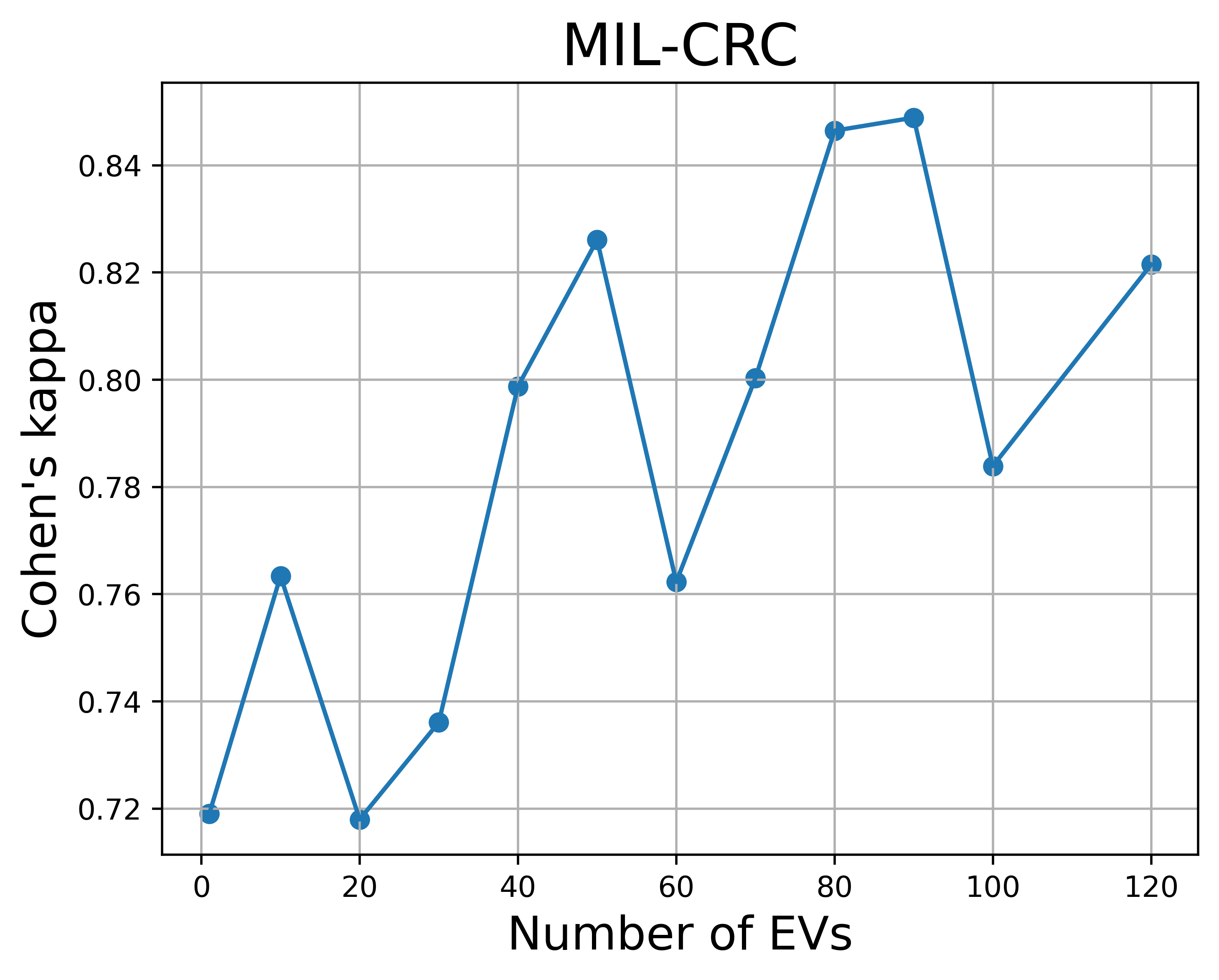}}
    \caption{Validation sets metrics as a function of the number of the selected eigenvectors (EVs), averaged over the 5-folds for varying numbers of eigenvectors. (a) F1-score, (b) Cohen's kappa score.} 
    
    \label{fig:ev_comparison}
    
\end{figure}

Figure~\ref{fig:ev_comparison} illustrates the average accuracy of our CIMIL-CRC model across the 5 validation set folds, plotted against the number of eigenvectors inputted into the final patient-level classifier.

Through our analysis, it was determined that using 90 eigenvectors produced the highest average F1-score and Cohen's kappa score across the validation sets. Furthermore, we identified that label smoothing rates of $\alpha=0.01$ or $\alpha=0.03$ resulted in the most favorable average F1-score and Cohen's kappa score.

\begin{figure}[t!]
    \centering
    \subfigure[]{\includegraphics[width=0.45\textwidth]{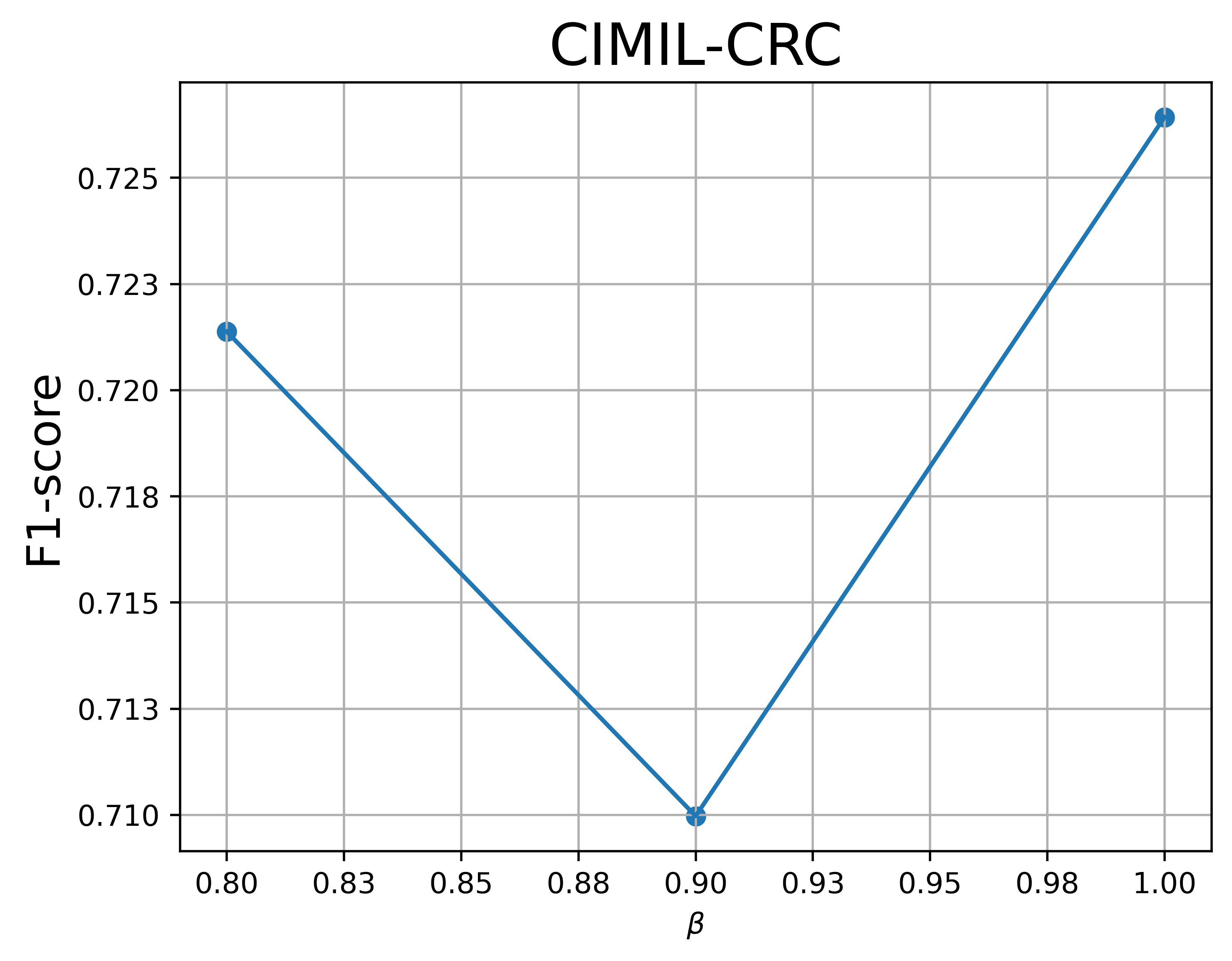}}
    \hfill
    \subfigure[]{\includegraphics[width=0.45\textwidth]{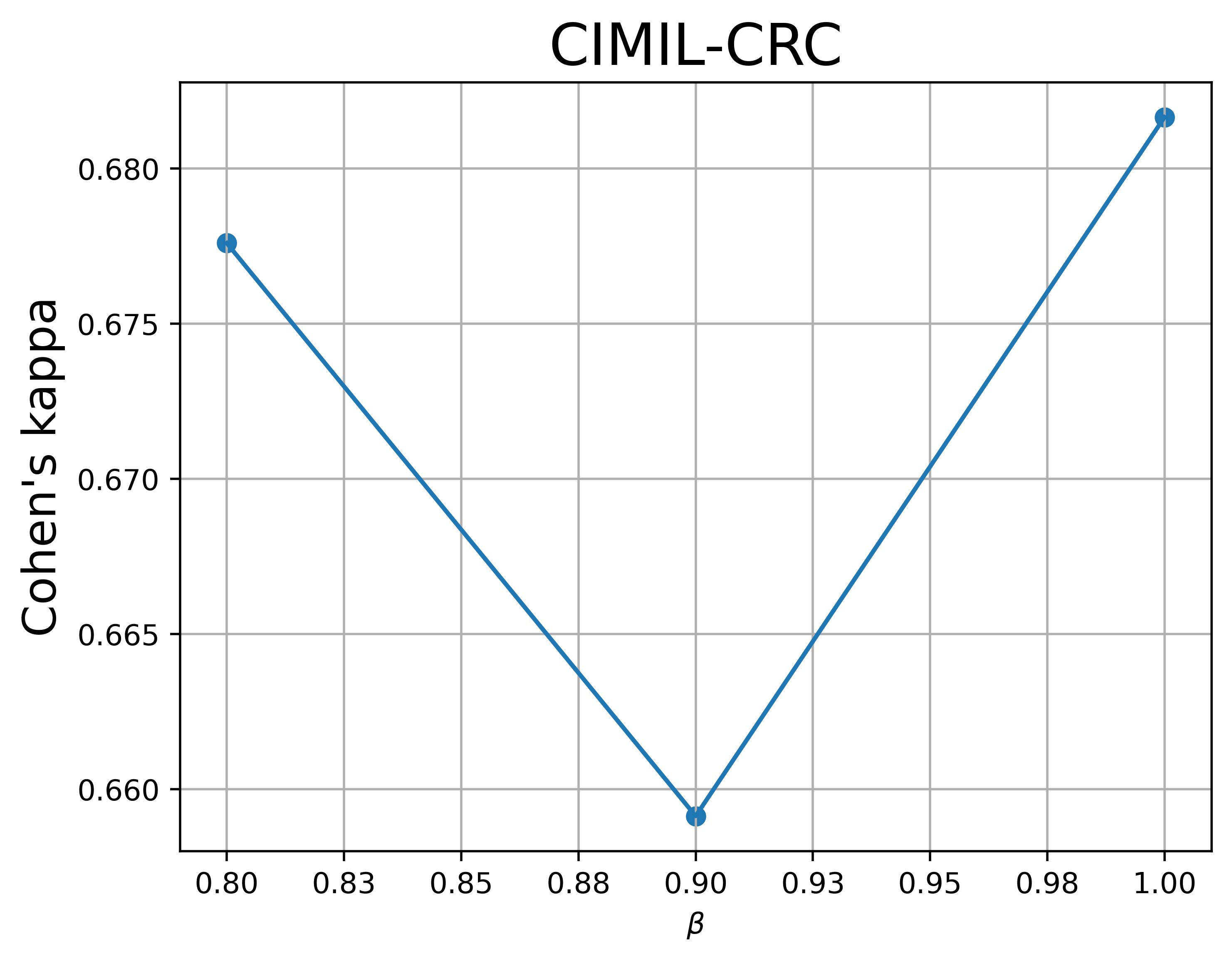}}
    \caption{Impact of the right side weight on the validation sets, averaged over the 5-folds for varying values of right side weight. (a) F1-score, (b) Cohen's kappa score.} 
    
    \label{fig:side_weight_comparison}
    
\end{figure}

We assessed the impact of the side weight $\beta$ for right-sided tumors or those with an undefined location on the validation sets by examining the following options: $\beta$ = [0.8, 0.9, 1].  Figure~\ref{fig:side_weight_comparison} depicts the impact of $\beta$ on the validation sets performance averaged over the 5 folds. We set $\beta$ to 1 as it achieved the highest F1-score and Cohen's kappa score.

These hyper-parameters were employed to train the models across various folds for our experimental analysis on the test set provided by Kather et al. \cite{Kather2019DeepCancer}.

\begin{figure*}
    \center
    \includegraphics[width=\textwidth]{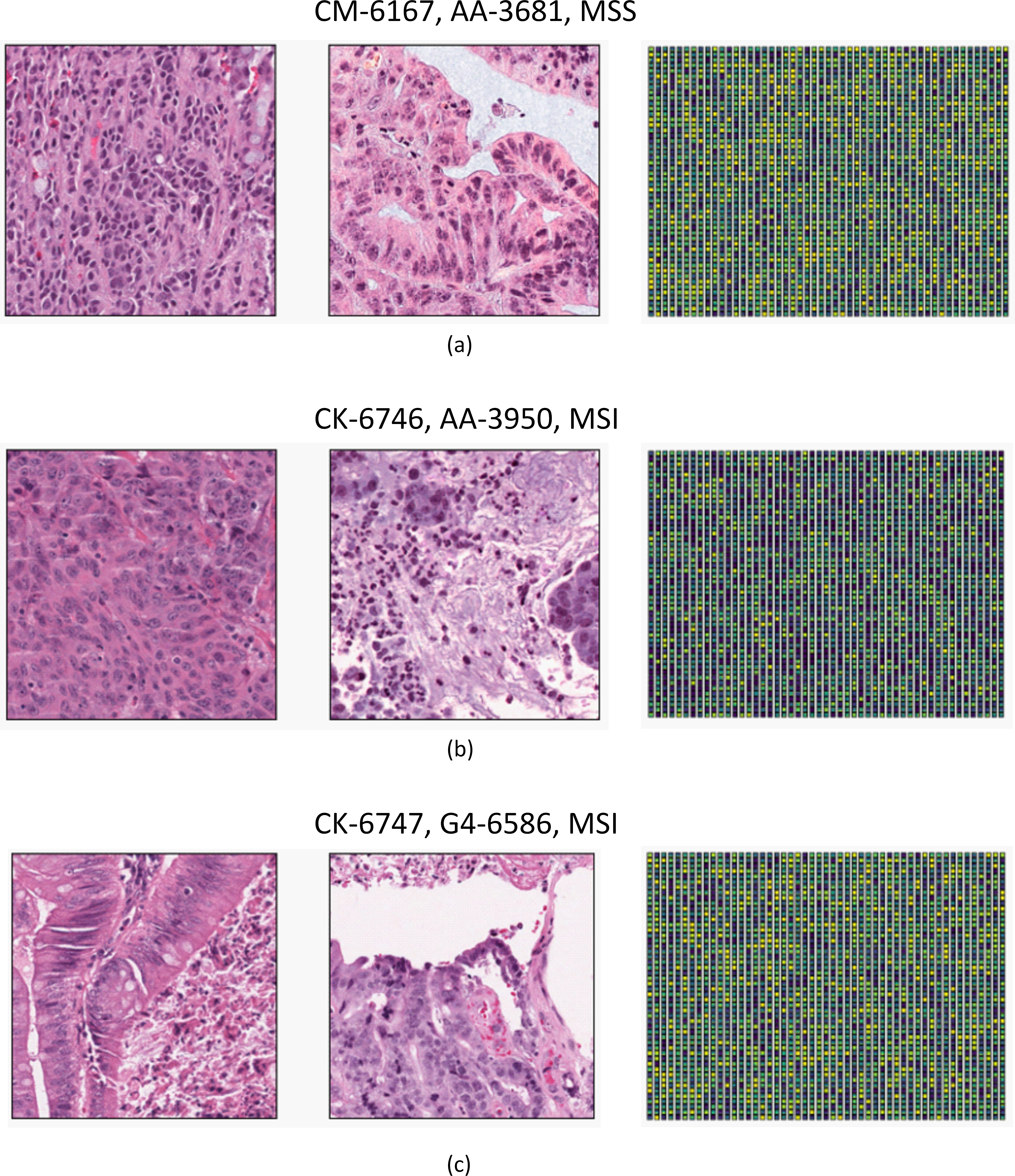} 
    \caption{Visualization of patches (first two columns) and corresponding representation of the first eigenvector \((7 \times 7 \times 7 \times 2560)\) of the leftmost patient (third column) from selected patients. The first and second rows show patients correctly classified by all models: Baseline patch classification, CI-Baseline, MIL-CRC, and CIMIL-CRC. The third row displays patches from patients which were correctly classified only by our CIMIL-CRC. }
    \label{fig:patches_classification}
\end{figure*}

Figure~\ref{fig:patches_classification} presents representative examples of patches from patients who were classified correctly as MSI or MSS (first two rows) by all models and a patient who was classified correctly as MSI only by our CIMIL-CRC model along with the first eigenvector used in our MIL-CRC and CIMIL-CRC models. 

\begin{figure}
    \center
    \includegraphics[width=\columnwidth]{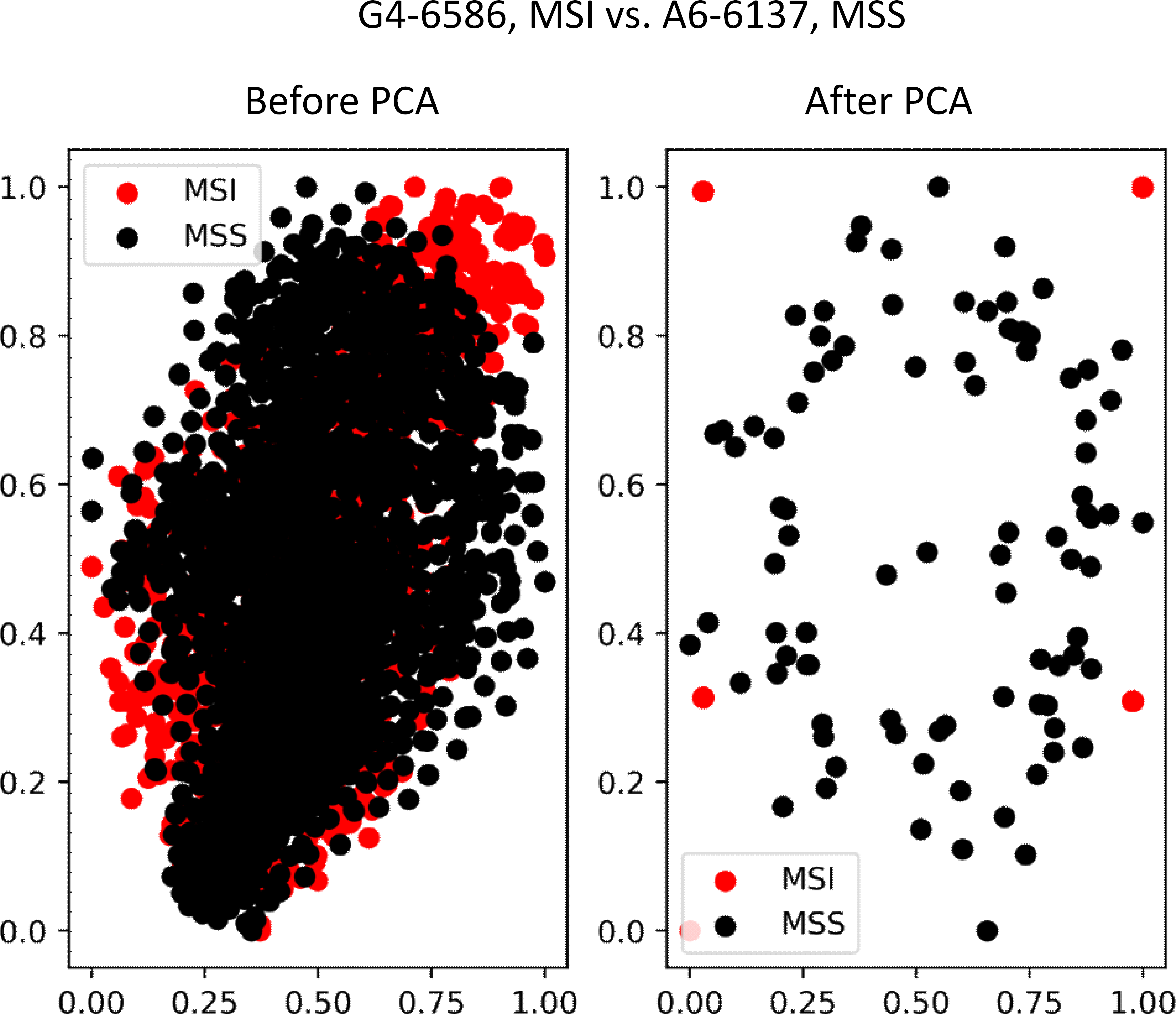} 
    \caption{The feature space, originally of dimensions \((7 \times 7 \times 2560)\), is visualized here as a 2D plot using t-SNE. Before applying PCA, each point represents an individual patch, whereas, after PCA, each point corresponds to an eigenvector. A comparison between an MSI and an MSS patient is plotted together to emphasize the distinction within their respective feature spaces. }
    \label{fig:pca_rep}
\end{figure}

Figure~\ref{fig:pca_rep} presents the t-SNE 2D representation of the feature space used as input to the MIL framework with and without applying PCA. Before applying PCA, each point represents an individual patch, whereas, after PCA, each point corresponds to an eigenvector. The application of PCA improves the separability between MSI and MSS patients.

Table~\ref{tab:our_results} presents a comparative analysis of our various models. The MIL-CRC model demonstrated a significantly higher average AUROC (0.86$\pm$0.01, 95\% CI 0.85-0.88, compared to 0.79$\pm$0.02, 95\% CI 0.76-0.82, with a paired t-test p-value$<$0.01) and AUPRC (0.8$\pm$0.02, 95\% CI 0.77-0.82, versus 0.64$\pm$0.04, 95\% CI 0.58-0.68, with a paired t-test p-value$<$0.01) compared to the baseline patch-level model. McNemar's test for the model's predictions was significant for all folds with p-value$<$0.03. Similarly, the CIMIL-CRC, which integrates clinical side information into our MIL framework, showed improved classification accuracy compared to the CI-Baseline, which is the clinical information integrated baseline patch-level model. This improvement is evidenced in both AUROC (0.92$\pm$0.002, 95\% CI 0.91-0.92, versus 0.88$\pm$0.007, 95\% CI 0.87-0.88, with a paired t-test p-value$<$0.001) and AUPRC (0.86$\pm$0.01, 95\% CI 0.84-0.87, compared to 0.81$\pm$0.01, 95\% CI 0.79-0.81, with a paired t-test p-value$<$0.01). McNemar's test for the model's predictions was significant for all folds with p-value$<$0.009.

\begin{figure*}[t!]
    \centering
    \subfigure[]{\includegraphics[width=0.45\textwidth]{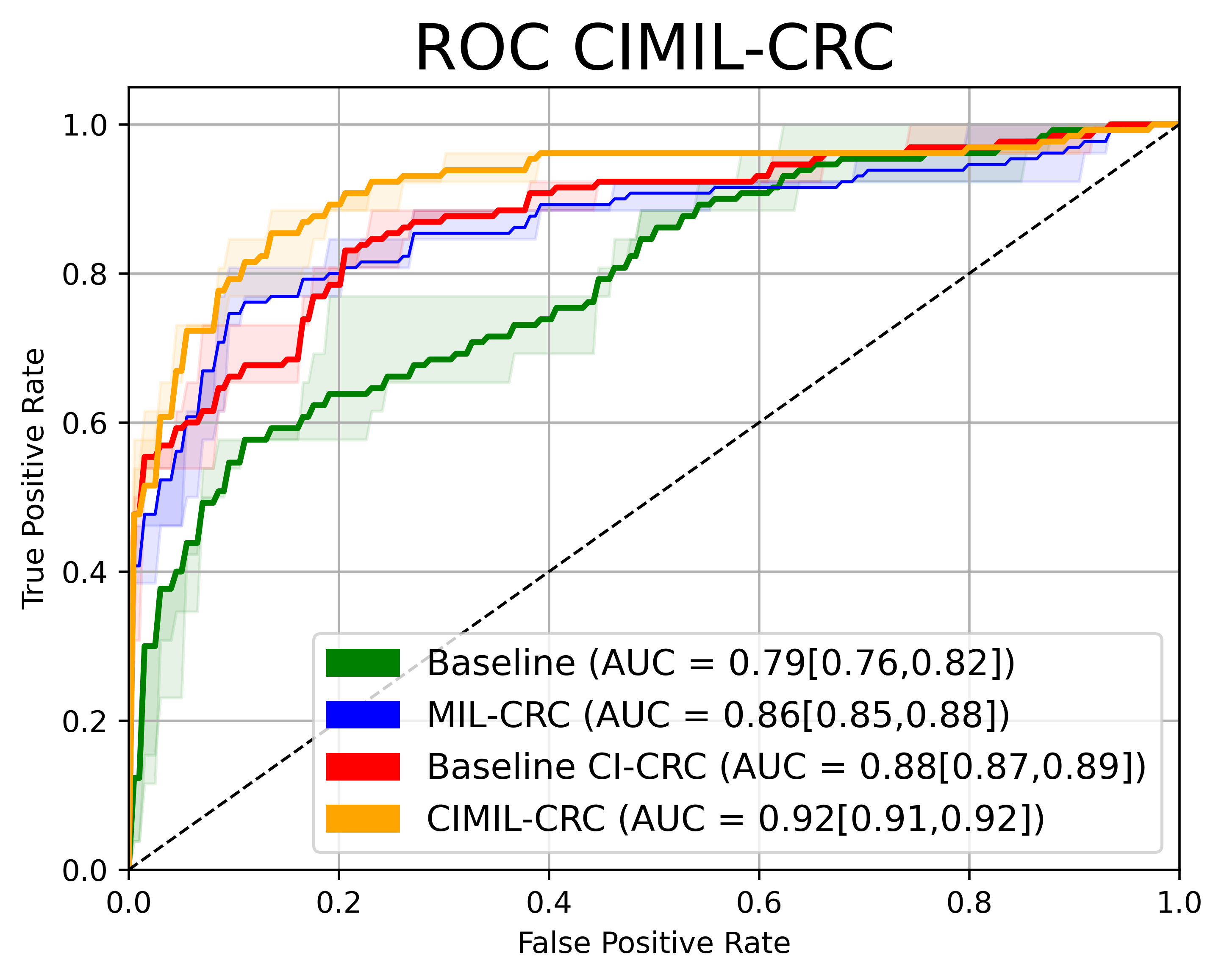}}
    \hfill
    \subfigure[]{\includegraphics[width=0.45\textwidth]{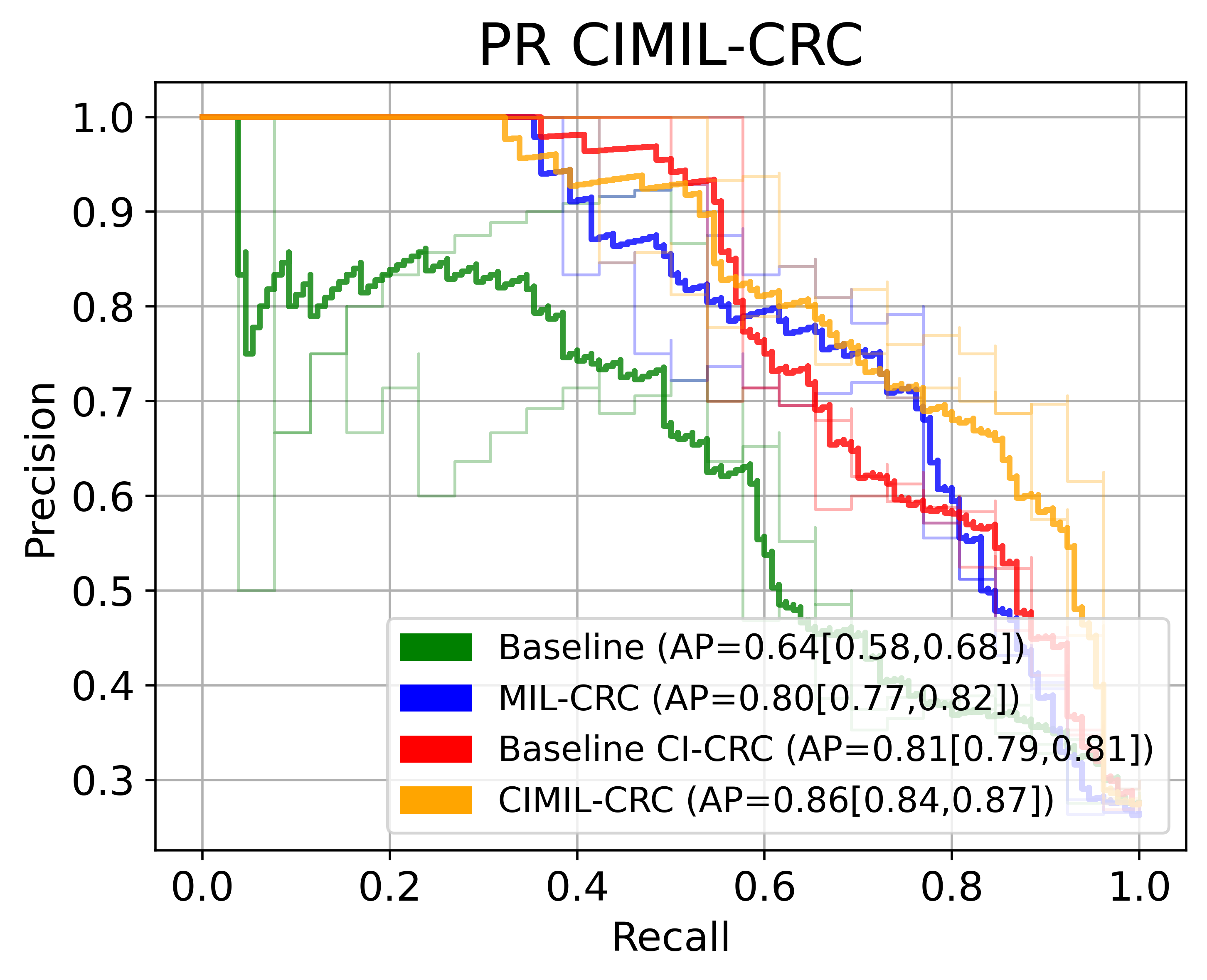}}
    \caption{Plots of MIL-CRC and CIMIL-CRC conducted with 90 eigenvectors plotted vs. the baseline patch classification and baseline with clinical side integration. (a) ROC curve (b) Precision-Recall curve.} 
    
    \label{fig:evmil_res}
    
\end{figure*}

Figure~\ref{fig:evmil_res} displays the ROC curves and the precision-recall curves for these different models on the test set provided by Kather et al. \cite{Kather2019DeepCancer}.

\begin{table*}[t]
 \caption{Performance of our MIL-CRC methods (MIL-CRC and CIMIL-CRC), averaged over 5 folds with 95\% confidence intervals, compared to the patch-classification methods (baseline and CI-Baseline). Note that CI-CRC does not have AUROC, AUPOC and standard deviation as it used a fixed formula, rather than parameters that depend on the different folds.}
\adjustbox{max width=\textwidth} {
   \centering
    \begin{tabular}{rrrrrr}
    %\begin{tabularx}{\textwidth}{rrrrrr}
    \hline
         Method & AUROC & AUPRC & F1-Score & Cohen's kappa & Accuracy\\
         \hline
         Baseline & 0.79(0.76-0.79) & 0.64(0.58-0.68) & 0.63(0.58-0.66) & 0.52(0.44-0.58) &0.82(0.8-0.86)  \\
         CI-Baseline & 0.88(0.87-0.88) & 0.8(0.79-0.81) & 0.72(0.69-0.76) & 0.63(0.56-0.68)& 0.88(0.87-0.89)\\
          CI-CRC & - & - & 0.71 & 0.4 & 0.7 \\
         MIL-CRC &  0.86(0.85-0.88) &  0.8(0.77-0.82) & 0.75(0.73-0.78)&0.67(0.64-0.71) & 0.87(0.86-0.89)\\
         \textbf{CIMIL-CRC} &  \textbf{0.92(0.91-0.92)} & \textbf{0.86(0.84-0.87)} & \textbf{0.8(0.79-0.82)} & \textbf{0.73(0.72-0.76)} & \textbf{0.89(0.88-0.91)}\\
         \hline
    \end{tabular}}
   
    \label{tab:our_results}
    %\end{adjustbox}
\end{table*}

  % \begin{figure}[t]
 
  %    \begin{tabular}{c}
  %       % \begin{adjustbox}{width=\linewidth} 
  %          \includegraphics[width=0.8\linewidth]{images/roc_CIMIL-CRC.png}\\
  %         (a)\\
  %         \includegraphics[width=0.8\linewidth]{images/ap_CIMIL-CRC.png}\\
  %         (b)\\
  %     %      \end{adjustbox}    
  %    \end{tabular}
  %   \caption{ }
  %     \label{fig:evmil_res}
  
  %   \end{figure}

Table~\ref{tab:ext_results_comp} showcases the performance of our CIMIL-CRC approach relative to recent state-of-the-art methods applied to the dataset from Kather et al. \cite{Kather2019DeepCancer}. Our method surpasses most others, particularly in terms of AUPRC, by a significant margin. It achieves an AUROC comparable to that of Saillard et al. \cite{saillard2021self} but with a notably smaller standard deviation. It's crucial to highlight that while Saillard et al.  \cite{saillard2021self} utilized additional data beyond the TCGA-CRC-DX dataset for training, our approach did not. Furthermore, we provide AUPRC and F1-score metrics, which are not reported by Saillard et al. \cite{saillard2021self}, making a direct comparison somewhat complex.

\begin{table}[t]
\caption{Performance of our MIL methods (MIL-CRC and CIMIL-CRC), averaged over 5 folds along with the std. over the folds, compared to reported published results (validated by their protocol). Note that previous works may not include the std., AUPRC, and/or F1-score. In these cases, we were not able to report these numbers in the table.}
\adjustbox{max width=\columnwidth} {
   \centering
   % \resizebox{\columnwidth}{!}{
    \begin{tabular}{lcrrr}
    \hline
         Method & AUROC & AUPRC & F1-score\\
         \hline 
         IDaRS \cite{Bilal2021DevelopmentStudy} & 0.9$\pm$0.01 & 0.72$\pm$0.02 &  -\\
         ResNet \cite{kather2020pan} & 0.84$\pm$0.01 & - & - \\
         RAMST \cite{lv2022joint} & 0.92 & - & - \\
         EPLA \cite{caodevelopment} & 0.89 & - & - \\
         simCLR+VarMIL \cite{schirris2022deepsmile} &0.86$\pm$0.02 & - & 0.47$\pm$0.1\\ 
         simCLR+patch classification \cite{schirris2022deepsmile} & 0.87$\pm$0.01 & - & 0.61$\pm$0.11\\ 
         MIL \cite{leiby2022attention} & 0.86$\pm$0.01 & 0.73$\pm$0.02 & -\\ 
         MIL+contrastive \cite{leiby2022attention} & 0.86$\pm$0.01 & 0.69$\pm$0.04 & -\\ 
         Moco+MIL \cite{saillard2021self} & 0.92$\pm$0.04 & - & - \\
         MIL-CRC(ours) &  0.86$\pm$0.01 & 0.8$\pm$0.02 &   0.75$\pm$0.02\\
         \textbf{CIMIL-CRC(ours)} &  \textbf{0.92$\pm$0.002} & \textbf{0.86$\pm$0.01} &  \textbf{0.8$\pm$0.01}\\
         \hline
    \end{tabular}
     }
    
    \label{tab:ext_results_comp}
\end{table}

\section{Discussion and Conclusion} 
We introduced the CIMIL-CRC framework, a clinically-informed MIL strategy for categorizing H\&E stained WSI of CRC into MSI and MSS subtypes. Contrary to previous MIL patient-level classification methods, which prioritized selecting representative patches and discarding less informative ones \cite{schirris2022deepsmile,Bilal2021DevelopmentStudy}, our approach generates a patient-level embedding space by applying PCA decomposition to patch-level features. Additionally, we incorporate clinical information, specifically tumor location, to improve classification accuracy. Our experiments, conducted in 5 folds on the TCGA-CRC-DX dataset \cite{crc-data,Kather2019DeepCancer}, have shown our approach's superiority over existing methods on the same dataset. Moreover, we have identified the distinct contributions of each aspect of our method, including the PCA-based MIL and the integration of clinical information.

It is important to note that selecting certain eigenvectors is not equivalent to patch selection. Unlike patch selection, where only specific patches are chosen, our approach involves transforming all patches of a patient into the eigenvectors' axis. We then select the eigenvectors with the highest explained variance. These eigenvectors encapsulate features derived from the entirety of the patient's patches, providing a comprehensive representation of the data.

Further, our proposed methodology not only emphasizes the critical role of patient-level aggregation in MIL-based classification but also highlights the significance of integrating prior knowledge and its effect on model performance. This approach shows promise for further application in other MIL-based tasks using H\&E WSI data.

While our model outperforms existing methods, there are inherent limitations that need addressing in future research. Our reliance on the TCGA-CRC dataset, pre-processed by Kather et al. \cite{Kather2019DeepCancer}, provides a robust platform for standardizing comparisons across different methodologies. However, this reliance also introduces certain constraints, notably in the form of potential biases inherent in the dataset and the limitations imposed by the specifics of the preprocessing steps used by Kather et al. \cite{Kather2019DeepCancer}. The issue of random patch sampling, especially in MSS patients, could introduce variability that might not be entirely representative of wider clinical realities.

In addition, the relatively small size of the dataset used in this study poses a significant limitation. Due to the rarity of datasets with the required molecular analysis, our findings may not fully capture the variability present in broader clinical settings. This limitation underscores the need for caution when generalizing our results and highlights the importance of future studies with larger, more diverse datasets to validate and extend our findings.

Further, the results achieved on the TCGA-CRC-DX cohort prompt considerations regarding the generalizability of the CIMIL-CRC framework to other datasets and broader clinical settings. While the integration of clinical priors enhances specificity, it also raises questions about the model's performance in environments where such detailed clinical data might not be readily available or may differ significantly. Additionally, future iterations of the CIMIL-CRC model could benefit from exploring alternative DNN architectures and more advanced feature extraction techniques that could potentially capture subtler nuances in histopathological images. Finally, expanding the framework to integrate other forms of clinical data, such as genetic markers and patient demographic information, might further enhance its diagnostic accuracy and clinical relevance.

In conclusion, the CIMIL-CRC method not only sets a new benchmark for MSI/MSS classification in colorectal cancer but also highlights the critical importance of integrating diverse data streams—ranging from detailed image analyses to nuanced clinical insights—into cancer diagnostics. The potential of this approach to refine the accuracy of cancer subtype classification and thereby inform more tailored treatment strategies represents a significant step forward in the personalized medicine landscape. Moving forward, addressing the outlined limitations and exploring broader applications of this technology will be crucial in realizing its full potential in clinical practice.

\section*{Acknowledgements}
M.F. acknowledges funding from the Israel Innovation Authority (grant number 73249) and from Microsoft Education and the Israel Inter-university Computation Center (IUCC). 
Y.E.M. acknowledges funding from the Israel Science Foundation (ISF, grant number 2794/21) and from the Israel Cancer Association (ICA, grant number 20210132).

\section*{Declaration of Generative AI}
During the preparation of this work, the author(s) used ChatGPT in order to improve readability. After using this tool/service, the author(s) reviewed and edited the content as needed and take(s) full responsibility for the content of the publication.

\bibliographystyle{plain}
% \biboptions{authoryear}
\bibliography{main}

\end{document}